\documentclass[10pt,twocolumn,letterpaper]{article}

\usepackage{caption}
\usepackage{iccv}
\usepackage{times}
\usepackage{epsfig}
\usepackage{graphicx}
\usepackage{amsmath}
\usepackage{amssymb}
\usepackage{gensymb}
\usepackage{bm}
\usepackage{verbatim}
\usepackage{multirow}
\usepackage{authblk}
\usepackage{multicol}

\DeclareMathOperator{\relu}{ReLU}

\usepackage[pagebackref=true,breaklinks=true,letterpaper=true,colorlinks,bookmarks=false]{hyperref}

\iccvfinalcopy %

\def\iccvPaperID{1422} %

\ificcvfinal\fi
\begin{document}

\title{Boundless: Generative Adversarial Networks for Image Extension}

\makeatletter
\renewcommand\Authfont{\fontsize{11.5}{14.4}\selectfont}
\renewcommand{\Authsep}{\qquad}
\renewcommand{\Authand}{\qquad}
\renewcommand{\Authands}{\qquad}
\makeatother
\author{Piotr Teterwak}
\author{Aaron Sarna}
\author{Dilip Krishnan}
\author{Aaron Maschinot}
\author{\\ \vspace{-10pt}David Belanger}
\author{Ce Liu}
\author{William T. Freeman}
\affil{Google Research \\
\tt \small \{pteterwak, sarna, dilipkay, amaschinot, dbelanger, celiu, wfreeman\}@google.com}

\maketitle
\ificcvfinal\thispagestyle{empty}\fi

\begin{abstract}

Image extension models have broad applications in image editing, computational photography and computer graphics. While image inpainting has been extensively studied in the literature, it is challenging to directly apply the state-of-the-art inpainting methods to image extension as they tend to generate blurry or repetitive pixels with inconsistent semantics. We introduce semantic conditioning to the discriminator of a generative adversarial network (GAN), and achieve strong results on image extension with coherent semantics and visually pleasing colors and textures. We also show promising results in extreme extensions, such as panorama generation.

\end{abstract}
\section{Introduction}

\begin{figure}[t]
    \includegraphics[width=.48\textwidth]{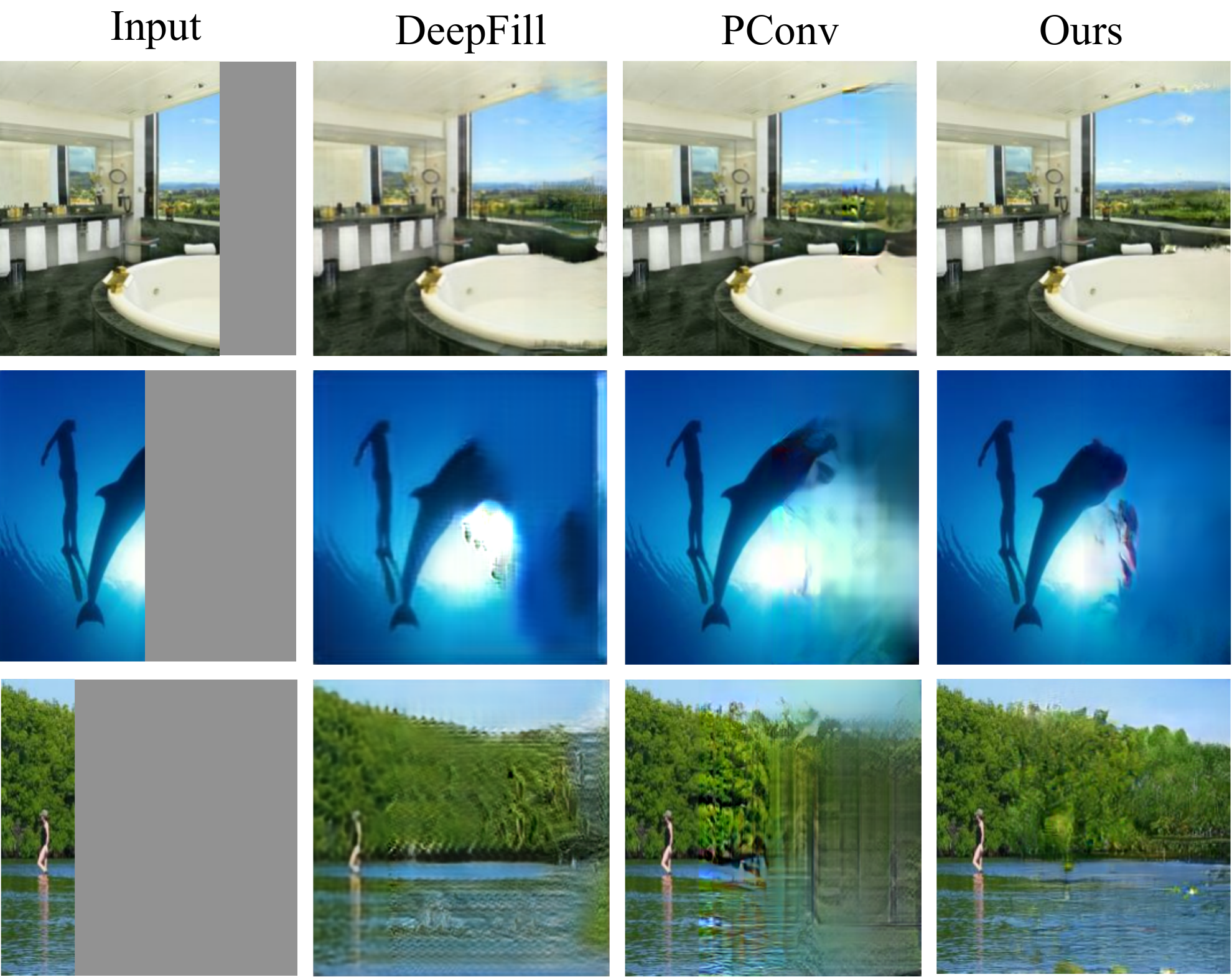}
    \captionof{figure}{Some examples of image extension: Our method (right column) generates better object shapes (top/middle rows) and produce good textures (middle/bottom rows), compared with two state of the art inpainting methods: DeepFill \cite{yu2018generative} and PConv \cite{liu2018partialinpainting}. The input image is extended onto the masked area (shown in gray).}
    \label{fig:teaser}
\end{figure}

Across many disparate disciplines there exists a strong need for high quality image extensions. In virtual reality, for example, it is often necessary to simulate different camera extrinsics than were actually used to capture an image, which generally requires filling in content outside of the original image bounds \cite{kaneva2010infinite}. Panorama stitching generally requires cropping the jagged edges of stitched projections to achieve a rectangular panorama, but high quality image extension could enable filling in the gaps instead \cite{kopf2012quality}. Similarly, extending videos has been shown to create more immersive experiences for viewers \cite{agarwala2005panoramic}. As televisions transition to the 16:9 HDTV aspect ratio, it is appealing to display videos filmed at a different aspect ratio than the screen. \cite{knee2010aspect, rubinstein2008improved}.

 We desire a seamless blending between the original and extended image regions. Moreover, the extended region should match the original at the textural, structural and semantic levels, while appearing a plausible extension. Boundary conditions are only available on one side of the extended region. This is in contrast to the image inpainting problem \cite{liu2018partialinpainting, yu2018generative}, where the region to be filled in is surrounded in all directions by original image data, significantly constraining the problem. Therefore, inpainting algorithms tend to have more predictable and higher quality results than image extension algorithms. In fact, we demonstrate in this paper that using inpainting algorithms with no modifications leads to poor results for image extension.

In the literature, image extension has been studied using both parametric and non-parametric methods \cite{wang2014biggerpicture, zhang2013framebreak, perez2003poisson, barnes2009patchmatch, avidan2007seam}. While these methods generally do a good job of blending the extended and original regions, they have significant drawbacks. They either require the use of a carefully chosen guide image from which patches are borrowed, or they mostly extend texture, without taking into account larger scale structure or the semantics of an image. These models are only applicable in a narrow range of use cases and cannot learn from a diverse data set. In practice, we would like image extension models that work on diverse data and can extend structure.

Fast progress in deep neural networks has brought the advent of powerful new classes of image generation models, the most prominent of which are generative adversarial networks (GANs) \cite{NIPS2014_5423} and variational autoencoders \cite{kingma2013auto}. GANs in particular have demonstrated the ability to generate high quality samples. In this paper, we use GANs, modified as described below, to learn plausible image extensions from large datasets of natural images using self-supervision, similar in spirit to the use of GANs in applications such as inpainting \cite{iizuka2017globally} and image superresolution \cite{ledig2017photo}.  

For the image extension problem, while state-of-the-art inpainting models \cite{yu2018generative, yu2018free} provide us a good starting point, we find that the results quickly degrade as we extend further from the image border.  We start by pruning the components that do not apply to our setting and then adopt some techniques from the broader study of GANs. Finally, we introduce a novel method, derived from \cite{miyato2018cgans}, of providing the model with semantic conditioning, that substantially improves the results. In summary, our contributions are:
\begin{enumerate}
    \item We are one of the first to use GAN's effectively to learn image extensions, and do so reliably for large extrapolations (up to 3 times the width of the original).
    \vspace{-8pt}
    \item We introduce a stabilization scheme for our training, based on using semantic information from a pre-trained deep network to modulate the behavior of the discriminator in a GAN. This stabilization scheme is useful for any adversarial model which has a ground truth sample for each generator input. 
    \vspace{-8pt}
    \item We show empirically that several architectural components are important for good image extension. We present ablation studies that show the effect of each of these components. 
\end{enumerate}

\section{Related Work}
\label{sec:related}
Prior work in image inpainting can be fairly neatly divided into two subcategories: classical methods, which use non-parametric computer vision and texture synthesis approaches to address the problem, and learning-based methods, which attack the problem using parametric machine learning, generally in the form of deep convolutional neural networks. Classical methods, such as \cite{ballester2001filling, bertalmio2000image, efros1999texture, efros2001image} typically rely on patch similarity and diffusion to borrow information from the known regions of the image to fill in the hole. These methods work best when inpainting small holes in stationary textures and generally lack semantic understanding of the image. Perhaps the most successful of these methods are the Bidirectional Similarity \cite{simakov2008summarizing} and PatchMatch algorithms \cite{barnes2009patchmatch,kopf2012quality}. Other non-parametric approaches that specifically target image extension rely on image patches from images other than the one to be extrapolated. \cite{shan2014photo, wang2014biggerpicture, sivic2008creating} rely on having large databases of photos available during the extrapolation process, while others, such as \cite{zhang2013framebreak}, depend on a carefully selected guide image.

In recent years, deep learning based approaches have made great strides in overcoming the weaknesses of the classical methods. The first significant learning-based approach to inpainting was the Context Encoder \cite{Pathak2016ContextEF}, which trained  an encoder-decoder model to fill in a central square hole in an image, using a combination of $\ell_2$ regression on pixel values, and an adversarial loss \cite{NIPS2014_5423}. \cite{yang2017high} minimizes the difference of nearest neighbor activation patches in deep layers of a pretrained ImageNet classification network, for improved synthesis of highly textured content. \cite{IizukaSIGGRAPH2017} improve on the results of \cite{Pathak2016ContextEF} by adding a local discriminator loss to the original global discriminator loss; the local discriminator focuses on the realism of the synthetic content, while the global discriminator encourages global semantic coherence. \cite{yu2018generative} improves on \cite{IizukaSIGGRAPH2017} further by introducing a coarse-to-fine approach. Their model has two chained encoder-decoder sections, the second of which contains a contextual attention layer, which learns the optimal locations in the unmasked regions from which the model should borrow texture patches. Other similar approaches include \cite{liu2018partialinpainting,wang2018image,yu2018free}, while \cite{yeh2017semantic} train an unconditioned GAN to generate complete images from the target distribution and perform an inference-time optimization to search for the latent code that would produce the closest match to the known pixels of the masked image. The only previous fully-parametric approach to image extension that we are aware of is \cite{van2016pixel}, which showed impressive results using an auto-regressive model to extend 32x32 pixel images, including the ability to output multiple plausible completions. These are, however, too small for practical applications. Concurrent to our work is \cite{wang2019srn}, which is similar to ours but does not condition the discriminator with pre-trained features.

\section{Model}
\label{sec:model}

\begin{figure*}[h]
    \def\svgwidth{\textwidth}
    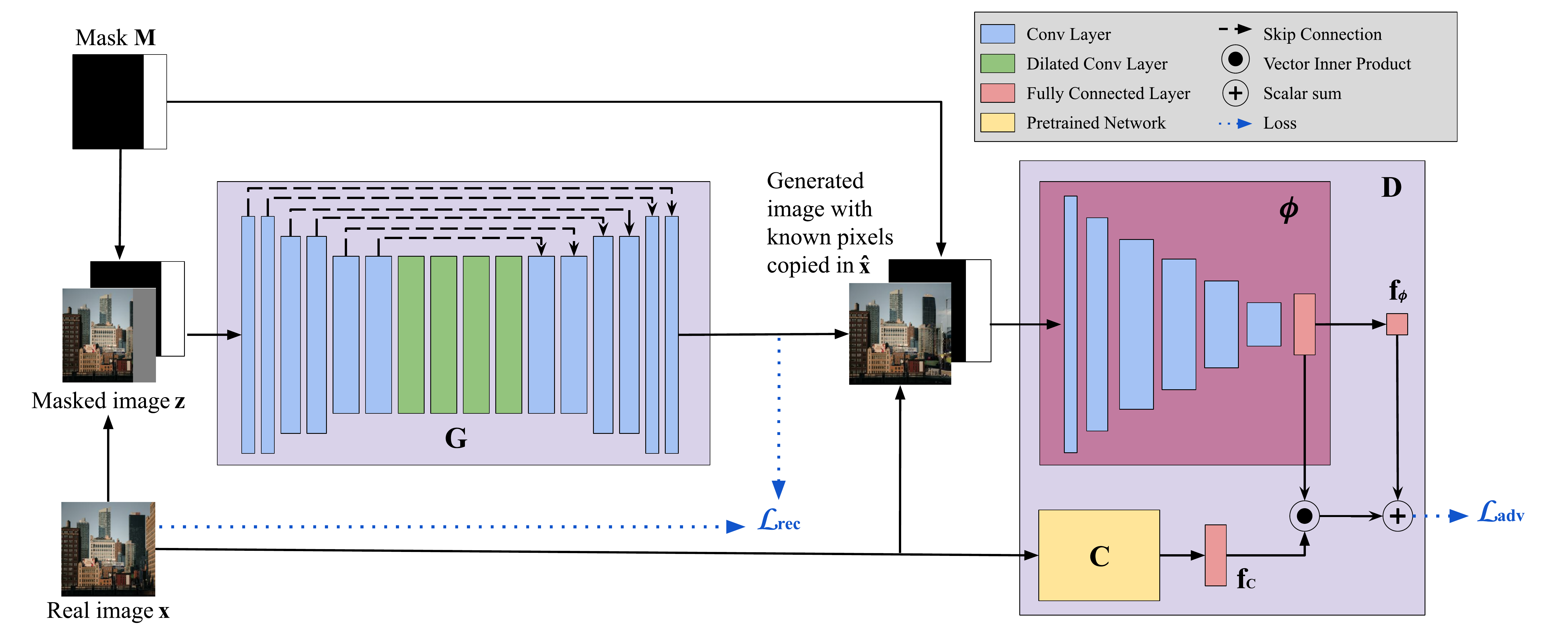
    \caption{Model Architecture: this architecture is used for all our models. See text for further details.}
    \label{fig:architecture}
\end{figure*}

Our model uses a Wasserstein GAN framework \cite{miyato2018spectral} comprising a generator network that is trained with the assistance of a concurrently-trained discriminator network. 

Our generator network, $\bm{G}$ has an input consisting of the image $\bm{z}$ with pixel values in the range $[-1, 1]$, which is to be extended, and a binary mask $\bm{M}$. These are the same dimensions spatially and are concatenated channel-wise. Both $\bm{z}$ and $\bm{M}$ consist of a region of known pixels and a region of unknown pixels. In contrast to inpainting frameworks, the unknown region shares a boundary with the known region on only \textit{one} side. $\bm{z}$ is set to $0$ in the unknown region, while $\bm{M}$ is set to $1$ in the unknown region and $0$ in the known region. At training time,
\begin{equation}
    \bm{z} = \bm{x} \odot (\bm{1} - \bm{M})
\end{equation}
where $\bm{x}$ is sampled from a natural image distribution $\bf{\mathcal{X}}$ and $\odot$ is the element-wise multiplication operator.

The output $\bm{G}(\bm{z}, \bm{M})$ of $\bm{G}$ has the same dimensions as $\bm{z}$ and a pixel loss during training uses this full output. However, the last stage before feeding into the discriminator $\bm{D}$ is to replace what $\bm{G}$ synthesized in the unmasked regions with the known input pixels: 

\begin{equation}
  \label{eq:xhat}
  \bm{\hat{x}} = \bm{G}(\bm{z}, \bm{M}) \odot \bm{M} + \bm{z}
\end{equation}

$\bm{D}$ is also a deep network, which transforms a real sample from $\bf{\mathcal{X}}$ or a generated sample $\bm{\hat{x}}$ to a single scalar value.

\subsection{Generator}

$\bm{G}$ generally follows the same fully convolutional encoder-decoder architecture as used by \cite{yu2018free} (see Figure \ref{fig:architecture}). Each layer in the generator except the last one uses gated convolutions \cite{yu2018free} to enable the model to learn to select the contributing features for each spatial location and channel. Following the inpainting guidance in \cite{yu2018generative}, each layer except the last uses an ELU activation function \cite{clevert2015fast}, and the final layer clips its outputs to the range $[-1, 1]$. As in \cite{iizuka2017globally, yu2018free, yu2018generative}, the innermost layers utilize dilated convolutions to increase their receptive field size.

\begin{figure}[h]
    \includegraphics[width=\columnwidth]{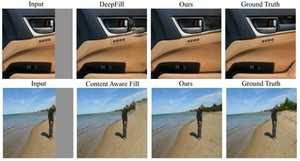}
    \caption{The contextual attention layer from  DeepFill \cite{yu2018generative, yu2018free} tends to repeat patches and structures. The original image (top left) is extended to the right (top-middle and top-right). DeepFill creates a copy of the door handle, whereas our extension extends the structure in a semantically and geometrically more plausible manner. Similarly, Photoshop's Content Aware Fill (bottom row) often creates artifacts since it is based on PatchMatch \cite{barnes2009patchmatch}.}
    \label{fig:compare_to_deepfill}
\end{figure}

To address the image extension problem, we deviated from the generator architecture proposed by \cite{yu2018free} in a few crucial ways. We eliminated the refinement network, including the contextual attention layer, since this layer is biased towards copying patches from the unmasked portion of the input. While borrowing patches is a useful property for inpainting of images \cite{barnes2009patchmatch}, in the case of image extension, it is less likely that repeated patterns will result in convincing extension. Figure \ref{fig:compare_to_deepfill} shows the effect of the contextual attention layer of \cite{yu2018generative, yu2018free}. We also compare to Adobe Photoshop's PatchMatch-based \cite{barnes2009patchmatch} Content Aware Fill tool, which generates similar artifacts due to copying patches. These copying artifacts occur on a large fraction of the output images.

We also introduced skip connections \cite{ronneberger2015u} between the non-dilated layers, since we found that they improved the network's ability to synthesize high frequency information. In Figure \ref{fig:inpainting_ablation}, we show the typical benefit of using skip connections. We additionally added instance normalization \cite{ulyanov2016instance} after every generator layer besides the output layer, finding that it significantly reduces the number of artifacts in the generated images.

\subsection{Discriminator}

The objective of the discriminator network (see Figure \ref{fig:architecture}) is determining whether an image is generator-produced or real. In our problem setup, the concern is not just whether the output of $\bm{G}$ appears real, but also that it is a plausible extension of $\bm{G}$'s inputs. To this end, we design our discriminator to be conditioned on the specific generator inputs when evaluating whether what is fed into the discriminator is real or fake. We condition the discriminator in two ways.

First, when a generated image is input, we copy the known pixels from $\bm{z}$ to overwrite the corresponding generated pixels, as described in eq. \ref{eq:xhat}, and we additionally input the mask $\bm{M}$ itself. This on its own provides a major advantage to the discriminator in the adversarial game, since it can focus in on the area right around the seam at the edge of the real content and easily determine that an image is fake if there is any abrupt change in image statistics along that seam. We see this play out during training, as the generated image content close to the seam is the first to improve and the quality improvement gradually spreads towards the opposite edge of the image as training progresses. On its own, this form of conditioning produces seamless results, but the quality of generated content still deteriorates as it moves further from the real content.

To address this, we add another form of conditioning, which is a modified version of the conditional projection discriminator (cGAN) \cite{miyato2018cgans}. In the original cGAN paper, a one-hot class label $\bm{y}$ is passed into the discriminator in addition to the image $\bm{x^*}$ to be classified as real or fake. The discriminator output is then 
\begin{equation}
  \bm{D}\left(\bm{x^*}, \bm{y}\right) = \bm{f_\phi}\left(\bm{\phi}\left(\bm{x^*}\right)\right)
                             + \left\langle \bm{\phi}\left(\bm{x^*}\right), \bm{f_y}\left(\bm{y}\right)\right\rangle
  \label{eq:cgan_discriminator}
\end{equation}
where $\bm{\phi}$ is a learned function mapping an image to a vector, $\bm{f_\phi}$ is a learned fully-connected layer that maps that vector to a scalar, $\bm{f_y}$ is a learned fully-connected layer mapping $\bm{y}$ to a vector of the same size as the output of $\bm{\phi}$, and $\left\langle \cdot, \cdot \right\rangle$ denotes an inner product. The cGAN paper shows that this parameterization of the GAN objective enables the model to simultaneously learn the distributions $p(\bm{x})$ and $p(\bm{y}|\bm{x})$.

In our setting we don't necessarily have class labels available, and we also want our conditioning vectors to contain more information than class labels would provide. To this end, we were inspired by previous work on perceptual metrics \cite{johnson2016perceptual, zhang2018unreasonable} to replace $\bm{y}$ with the activations of a pretrained image classification network, $\bm{C}$, when applied to $\bm{x}$ (the ground truth image). We chose to instantiate $\bm{C}$ as an InceptionV3 \cite{szegedy2016rethinking} network trained on ImageNet \cite{deng2009imagenet} with the final softmax removed. We found that it helps to normalize these activations by subtracting the mean activation over the dataset and then dividing the result by its $\ell_2$ norm. Note that since the discriminator is only used during training, we can condition on the full unmasked image ($\bm{x}$), which also means that these activations can be precomputed before training. This conditioning encourages the generated content to semantically match the target image, which especially helps avoid semantic drift in larger extensions. Formally, we replace eq. \ref{eq:cgan_discriminator} with
\begin{equation}
\begin{split}
  \bm{D}\left(\bm{x^*}, \bm{M}, \bm{x}\right) = & \bm{f_\phi}\left(\bm{\phi}\left(\bm{x^*}, \bm{M}\right)\right)\\
                            & + \left\langle \bm{\phi}\left(\bm{x^*}, \bm{M}\right), \bm{f_C}\left(\bm{C}\left(\bm{x}\right)\right)\right\rangle
  \label{model_discriminator}
\end{split}
\end{equation}

The architecture of $\bm{\phi}$ is based on \cite{yu2018free} and consists of six strided convolutional layers, followed by a fully connected layer. Each convolutional layer uses a leaky ReLU activation function \cite{maas2013rectifier} and all layers apply spectral normalization \cite{miyato2018spectral} to satisfy the Lipschitz constraints of Wasserstein GANs \cite{arjovsky2017wasserstein}. The output dimensions of $\bm{\phi}$ and $\bm{f_C}$ are both 256.

\subsection{Training}
\label{sec:Training}

The model is trained via a combination of a reconstruction loss and an adversarial loss. The reconstruction loss optimizes for coarse image agreement and is implemented as an $\ell_1$ loss imposed on the full output of $\bm{G}$.  The full equation is below:
\begin{equation}
  \mathcal{L}_{rec} = \left\Vert\bm{x} - \bm{G}\left(\bm{z}, \bm{M} \right)\right\Vert_1
\end{equation}
For the adversarial loss, which refines the coarse prediction, we use a Wasserstein GAN hinge loss \cite{lim2017geometric, tran2017deep}:
\begin{align}
  \begin{split}
    \mathcal{L}_{adv,D} = & \
    \mathbb{E}_{x \sim P_{\mathcal{X}}(x)} [ \
      \relu\left(1-\bm{D}\left(\bm{x},\bm{M}, \bm{x}\right)\right) + \\
      & \qquad \qquad \;\; \relu\left(1+\bm{D}\left(\bm{\hat{x}},\bm{M}, \bm{x}\right)\right) \
    ]
    \\
    \mathcal{L}_{adv,G} = & \
    \mathbb{E}_{x\sim P_{\mathcal{X}}(x)}[ \
      -\bm{D}\left(\bm{\hat{x}},\bm{M}, \bm{x}\right) \
    ]
  \end{split}
\end{align}
where $\relu$ is the rectified linear unit function. The total loss on the generator is
\vspace{-7pt}
\begin{equation}
    \mathcal{L}_{total} = \mathcal{L}_{rec} + \lambda \mathcal{L}_{adv,G}
    \vspace{-7pt}
\end{equation}
In all our experiments we set $\lambda = 10^{-2}$.

Our model is implemented in TensorFlow \cite{abadi2016tensorflow}. The generator and discriminator are trained jointly using the Adam optimizer \cite{kingma2014adam} with parameters $\alpha = 10^{-4}, \beta_1 = 0.5, \beta_2 = 0.9$; (the discriminator has a slightly larger $\alpha = 10^{-3}$, but other parameters are the same). Unlike many previous papers, we did not see improvement from training the discriminator for multiple steps per each generator step. Based on the findings of \cite{brock2018large} on the benefits of training GANs with large batch sizes, we trained on 32 cores of a Google Cloud TPUv3 Pod with batch size 256.

\section{Experimental Results}
\label{sec:results}
For all experiments we train our model on a dataset composed of the top 50 classes (measured by number of samples in the training set) of the Places365-Challenge dataset \cite{zhou2017places}, producing a training set of just under 2 million images, which we scaled to the spatial dimensions of 257x257 pixels. The use of 50 classes allows us to test how well our model can generalize to multiple categories. We used a held-out set of 500 images from the same set of classes in the Places365 dataset, approximately 10 images per class, to compute quantitative scores and to visualize the image extension results.

\subsection{Image Extension}
\label{subsec:extension}

\begin{table}[tp]
\small
\setlength{\tabcolsep}{2pt}
\centering
\begin{tabular}{|c|c|c|c|c|c|c|c|c|c|}
\hline
  & FID & PSNR & FID & PSNR & FID & PSNR & FID & PSNR\\
& (25\%) & (25\%) & (50\%) & (50\%) & (75\%) & (75\%) & (Inp) & (Inp) \\
\hline
\hline
DF & 1.87 & 7.11 & 11.65 & 6.69 & 31.21 & \textbf{9.74} & 4.96 & \textbf{14.31} \\
PC & 1.40 & \textbf{11.10} & 11.20 & 6.63 & 31.83 & 8.94 & 3.70 & 13.78 \\
NCnd & 0.85 &  8.96 & 5.01 & 7.55 & 19.17 & 9.08 & 2.73 & 14.24 \\
Ours & \textbf{0.79} & 10.17 & \textbf{3.46} & \textbf{8.63} & \textbf{8.79} & 8.07 & \textbf{2.53} & 14.17 \\
\hline
\end{tabular}
\vspace{4pt}
\caption{Quantitative metrics on $500$ test images.The mask types are: 25\% extension (3:1 ratio of context to mask), 50\% extension (1:1 ratio), 75\% (1:3 ratio) and inpainting a central square mask comprising 25\% of image pixels(Inp). We compare DeepFill(DF), PartialConv(PC), ours without conditioning(NCnd), and our model.}
\label{exp:table_all}
\end{table}

\begin{table}[tp]
\small
\setlength{\tabcolsep}{2pt}
\centering
\begin{tabular}{|c|c|c|c|c|c|c|c|c|c|}
\hline
  & FID & PSNR & FID & PSNR & FID & PSNR \\
& (25\%) & (25\%) & (50\%) & (50\%) & (75\%) & (75\%) \\
\hline
\hline
Prcptl & \textbf{0.40} &  9.95 & \textbf{2.32} & 8.31 & 14.15 & \textbf{9.65}  \\
FM & 0.75 &  9.42 & 3.14 & \textbf{8.97} & 14.74 & 8.87  \\
Ours & 0.79 & \textbf{10.17} & 3.46 & 8.63 & \textbf{8.79} & 8.07 \\
\hline
\end{tabular}
\vspace{4pt}
\caption{Comparisons with other methods for stabilizing GAN training. We provide PSNR for reference, but found that FID correlates with perceptual quality best. Based on FID on 25\% and 50\% extensions, feature matching and perceptual losses outperform our conditioning, but the difference is fairly small. On 75\% extensions, our conditioning provides the best results and the difference is large. }
\label{exp:table_fm_prcptl}
\end{table}

\begin{figure*}
	\centering
    \hspace{-10pt}\includegraphics[width=1.02\textwidth]{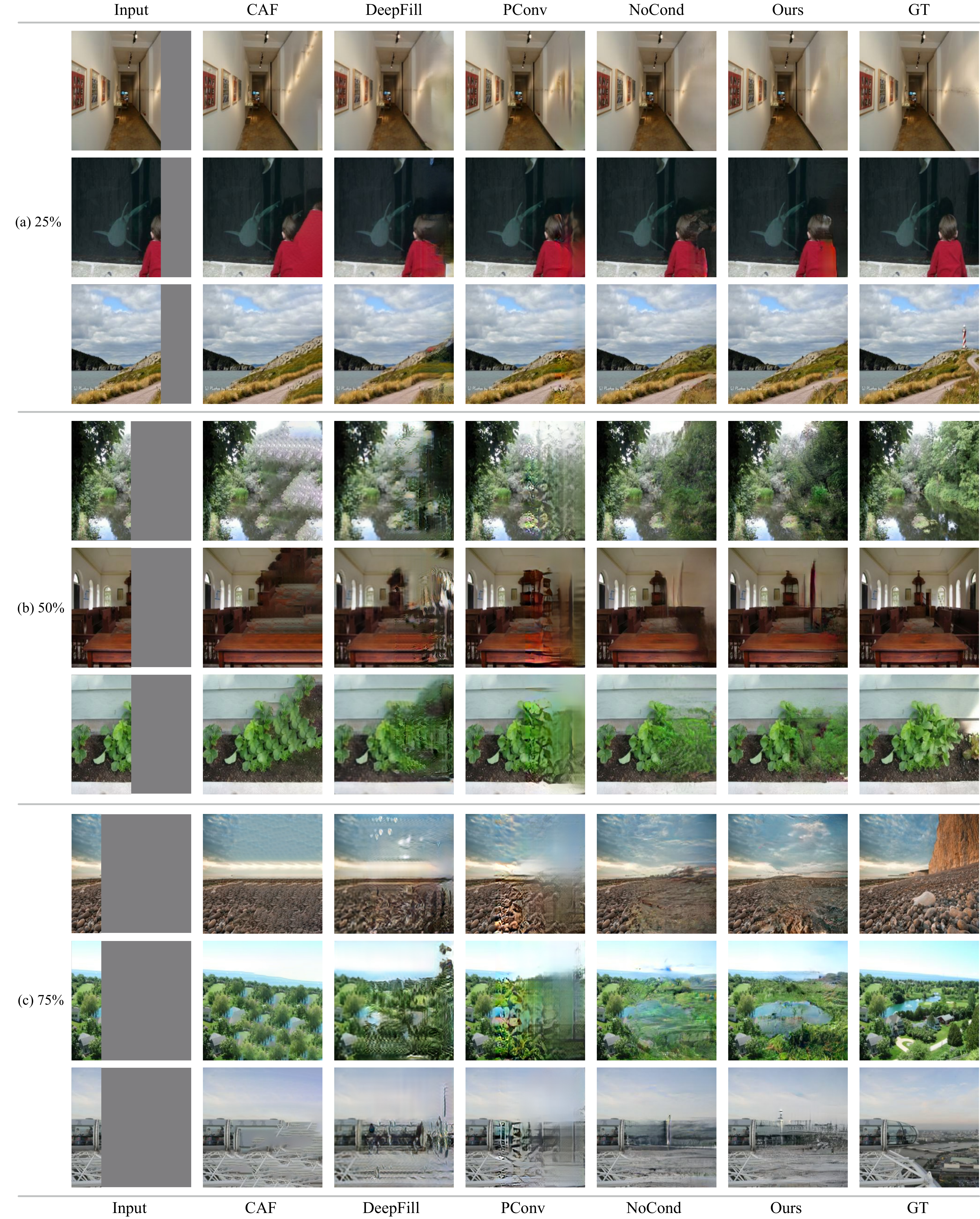}
    \caption{Extending images from masks of  (a) 25\%,  (b) 50\% and (c) 75\% of the image width using multiple algorithms. From left to right: DeepFill \cite{yu2018generative}, PConv \cite{liu2018partialinpainting}, Photoshop Content Aware Fill, our model with no conditioning, our full model and ground truth.}    
    \label{fig:uncrop_various_masks}
\end{figure*}

We compare our model (which we call \textbf{Ours}) with various baselines both qualitatively and quantitatively on the task of image extension. Specifically, we evaluate each algorithm's ability to fill in masked out image content for three different image extension tasks (where the rightmost 25\%, 50\% and 75\% of pixels in the image are respectively masked) and one inpainting task (a central square mask comprising 25\% of image pixels). For each experiment, our model is retrained with masks of the appropriate position, shape and size. The baselines we compare against are:\\
    \noindent\textbf{No-Cond}: A model that is identical to ``Ours," but without discriminator conditioning.\\
    \noindent\textbf{DeepFill}: Our re-implementation of  DeepFillv2 \cite{yu2018free}, which is a state of the art inpainting model. We confirmed that our reimplementation achieves inpainting results nearly identical to that of the original papers (see Figure \ref{fig:inpainting_ablation}). For each experiment, we retrain the model with the same masks and data as ours. We follow the authors' guidance of training for 5 days on an NVIDIA P100 GPU. We note that this results in the model being trained for many fewer steps than ``Ours" because it trains much more slowly (0.8 steps/sec vs 4.7 steps/sec for ``Ours"). Comparison against this model shows the benefits of our approach compared to simply repurposing an architecture suited for inpainting tasks.\\
    \noindent\textbf{PConv}: The authors of another state of the art inpainting work \cite{liu2018partialinpainting} generated results for us based on provided masks, but the models were not retrained specifically for these tasks. The model was trained on the full Places2 dataset, which is a superset of our training set. They use a database of free-form masks, some of which are very large (up to $50\%$ of the image size), but are often non-contiguous and non-convex, which means that at training time the model may not have needed to generate pixels that were very far from known context. While our comparisons to this paper are not exactly apples-to-apples, we believe that this still provides a strong baseline against which to compare our performance.\\
    \noindent\textbf{CAF}: Results from Adobe Photoshop's content aware fill, which is based on the PatchMatch  algorithm \cite{barnes2009patchmatch}. Content aware fill is a very powerful tool used for image extension, and reprents a strong classical baseline. However, due to the use of only patch level information, it does not provide semantically meaningful extensions.\\   
    
We provide quantitative performance metrics for each mask-type and each algorithm in Table \ref{exp:table_all}. We agree with the authors of \cite{yu2018generative, liu2018partialinpainting} that there are really no good metrics that capture the goals of these experiments, but we nonetheless report the best approximations. Specifically, we report Fr\'echet Inception Distance (FID) \cite{heusel2017gans} on the full output image and PSNR of the masked regions only. For FID we used a diagonal covariance matrix, due to having few samples. Based on our own qualitative evaluations, we feel that FID of the entire output image best correlates with what we perceive as quality image extension. %
We additionally performed a qualitative analysis. We show results on a few images from our test set in Figures \ref{fig:uncrop_various_masks} and \ref{fig:inpainting_ablation}. We show many more results, including on free-form masks, in the Supplementary Material. Overall we see that all methods, other than ``CAF," perform admirably for inpainting. On the extension tasks, as we move towards larger extensions with smaller context, ``CAF" and ``DeepFill" degrade into just repeating textures, while ``PConv" gets blurrier and more artifact-filled the further away from the context it gets. The ``NoCond" version of our model maintains higher quality for the larger extensions but does show some blurring and semantic drift. Meanwhile, our full model remains semantically consistent, with mostly photorealistic and seamless synthesis.

\begin{figure}[ht!]
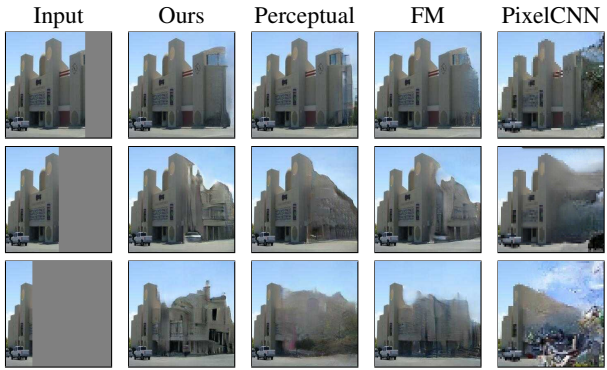

\newcommand{\figwidth}{.08\textwidth}
\newcommand{\shiftleft}{\hspace{-20pt}}
\newcommand{\figurefile}[2]{\frame{\includegraphics[width=\figwidth]{figs/perceptual_featurematch_comp/#1/#2}}}
\newcommand{\onerow}[1]{
\figurefile{#1}{input.jpg} & 
\figurefile{#1}{bndls.jpg} & 
\figurefile{#1}{prcptl.jpg} & 
\figurefile{#1}{fm.jpg} &
\figurefile{#1}{pixelcnn.jpg} 
}

\setlength{\tabcolsep}{1.2mm}
\begin{tabular}{ccccc}
 {\small Input}  &  {\small Ours} & {\small Perceptual}  &  {\small FM} & {\small PixelCNN} \\
\onerow{example_6}\\
\onerow{example_7}\\
\onerow{example_1}\\
\end{tabular}
\caption{\small Although FM and Perceptual give slightly better results on short range extensions, our model far outperforms the others in the 75\% case. }
\label{fig:perceptual}
\end{figure}

Furthermore, we experiment with replacing our conditioning with perceptual \cite{johnson2016perceptual} and feature matching \cite{wang2018pix2pixHD} losses, see Table \ref{exp:table_fm_prcptl} and Figure \ref{fig:perceptual}. In the perceptual loss, the generator is optimized to produce images that are close to the ground truth images in the activation space of a pretrained classification network. Similar to our conditioning, perceptual loss uses a pre-trained network to guide the training towards plausible extensions. Unlike our conditioning, the perceptual loss  modifies  only the generator objective to bias it towards semantic coherence, whereas our conditioning figures into both the generator and discriminator objectives and adding semantic information to the whole adversarial optimization game. 
Feature matching is similar in principle to a perceptual loss, but it minimizes the distance between activations in a hidden layer of the discriminator, rather than a pre-trained classification network, and similarly only figures into the generator objective.

In our experiments on 75\% extensions, we found that our conditioning performs significantly better than feature matching and perceptual loss, while on smaller extensions they perform similarly. Our perceptual loss implementation tries to match pre-softmax logits, while our feature matching follows \cite{wang2018pix2pixHD} and tries to match convolutional feature maps in the discriminator at multiple scales. Preliminary experiments indicate that combinations of all the three result in even better results, chiefly with fewer GAN-style artifacts, but we leave a more thorough analysis to future work.

We also qualitatively compare against PixelCNN in Figure \ref{fig:perceptual}. It is clear PixelCNN performs much worse than our method. Furthermore, it takes about 12 minutes to do inference for a single 64x64x3 image, on a Tesla P100. On our higher resolution 256x256 images, this would translate to over 2.5 hours. In contrast, our method takes 0.1 seconds/image.

\subsection{Ablations}
\label{subsec:ablation}

In order to validate the contributions of each aspect of our model, we trained models each ablating a single feature of our network. We experimented with removing skip connections in the generator, removing feature conditioning from the discriminator, removing instance norm from the generator and reducing batch size to 64.

\begin{figure*}
\newcommand{\figurewidth}{\textwidth}
\small
    \centering
    \begin{minipage}{\textwidth}
	    \centering
        \includegraphics[width=\figurewidth]{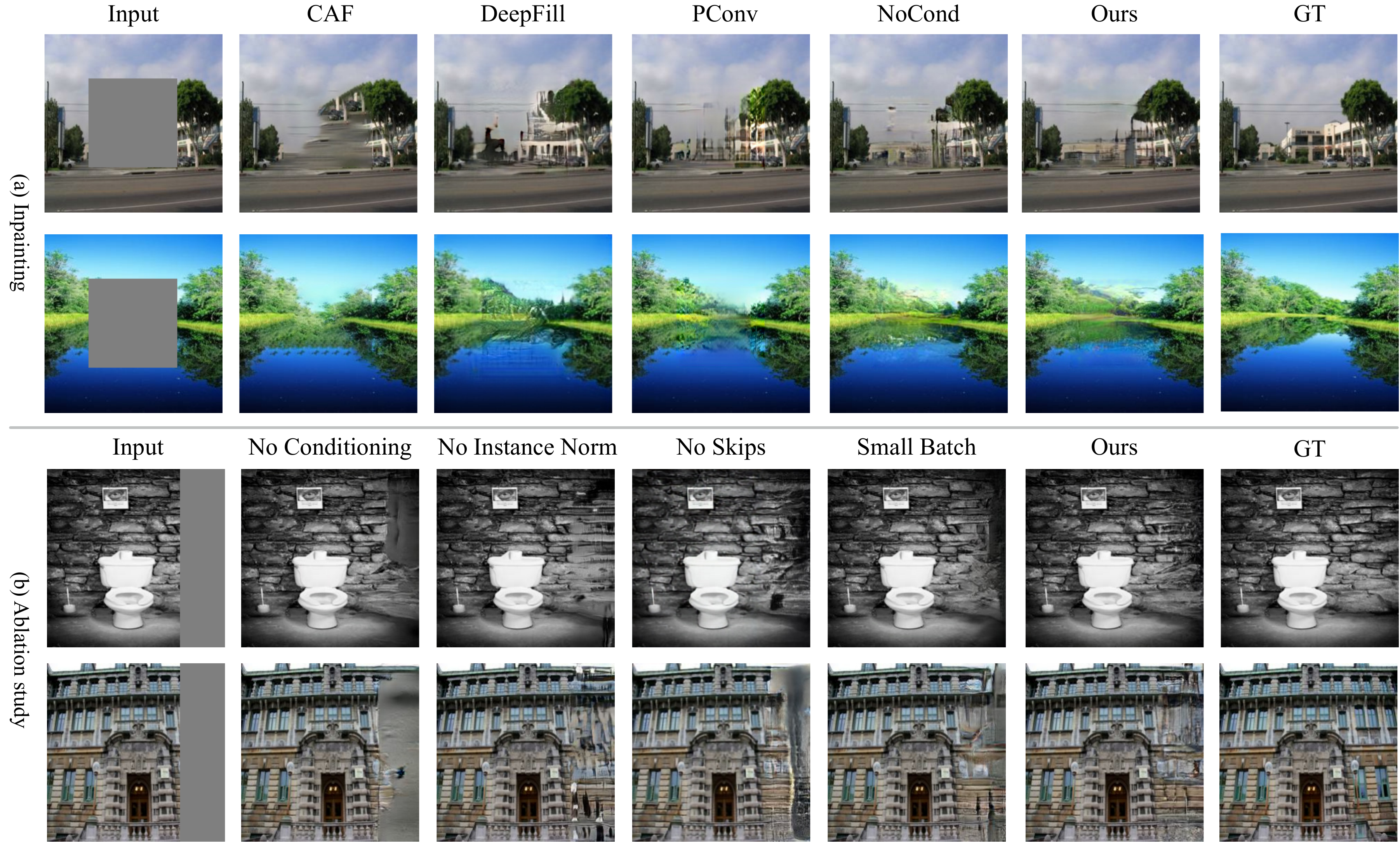}
        \caption{Further analysis: (a) Comparing the different models on inpainting tasks; our conditioned model performs on par with the state of the art models such as PConv and DeepFill for the inpainting problem. (b) Ablation tests: we remove (from second column to fifth column) only one of the following: discriminator conditioning, instance norm, skip connections, and reduce batch size.}    
        \label{fig:inpainting_ablation}
    \end{minipage}\\[5pt]
 
    \begin{minipage}{\textwidth}
    	\centering
        \includegraphics[width=\figurewidth]{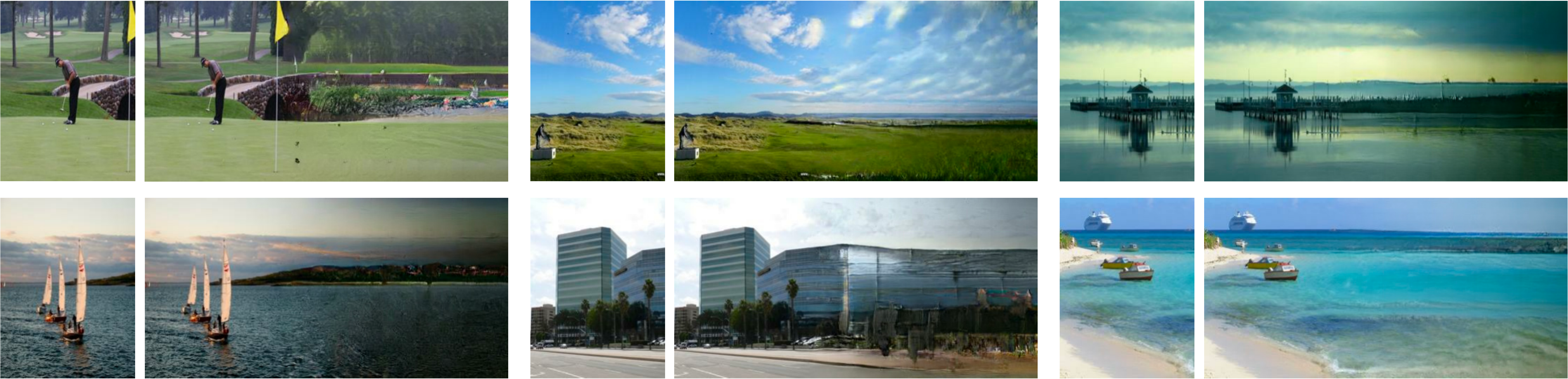}
        \caption{Our models can also be used generate image panoramas. This can be viewed as a stress test for image extension tasks. We recursively apply the 25\% model to create a very large output image of about 3 times the original width.}
        \label{fig:application:panorama}
    \end{minipage}    
\end{figure*}

We show an example of the results in Figure \ref{fig:inpainting_ablation}. We see that without skip connections, the model has a hard time synthesizing high frequency textures, which often results in a perceptible seam between the real and generated content. Without discriminator conditioning the quality degrades more rapidly as we move further away from the edge with known context. Removing instance norm causes an increase in white and over-saturated artifacts. The smaller batch model generally produces blurrier results.

\subsection{Panorama Generation}
\label{subsec:panorama}

We evaluate our model on its ability to generate panoramic images from a much narrower seed image. We use the same dataset and models as in Section \ref{subsec:extension}, but whereas in those experiments we ran each model's forward pass only once, this time we apply the model recursively, using the synthesized content from the previous step as the known content for the current step in a sliding window approach. More specifically, we take a 257x192 image, pad it with 65 columns of zeros, and extend it using our our model trained for 25\% extension. We then extend the image again by selecting the right most 192 pixels of the image, padding with zeros, and feeding it to the extension model. This is repeated 5 times, ultimately extending the image 2.7 times out to a width of 582 pixels. We show some results in Figure \ref{fig:application:panorama} and more results in the Supplementary Material. While the results are for the most part plausible and aesthetically pleasing, we do see some degradation and semantic drift as we move away from the original image. 

\subsection{Exploring the Space of Plausible Extensions}
\label{subsec:video}

To visually explore the space of results generated by our extension model, we took sample videos from the YouTube8m dataset \cite{abu2016youtube} and applied our model to extend both the right and left sides horizontally. Videos are a natural domain to test our model since consecutive frames are closely related to each other, and the small variations can generate interesting plausible outcomes. This allows us to verify that our model has not memorized a fixed completion for closely related images or collapses with small natural variations in the input. We scaled the videos to have a height of 257 pixels and respectively selected the right and leftmost 192 columns. We then padded each of them with 65 columns of zeros on the side to be extended. We ran the resulting images, each frame independently, through our model trained with 25\% masks. We then took the model output for the extended region and concatenated it back on the original video frame. Please refer to the supplementary video \url{https://drive.google.com/open?id=1x6FCYPmoqSuCdeLJTD0UpQ_MQhBPv7_e}.

\section{Acknowledgements}
The authors would like to acknowledge Amol Kapoor and Dr. Huiwen Chang for helpful discussion and sharing code, and Dr. Guilin Liu for helping with running Partial Convolution comparison experiments.

{\small
\bibliographystyle{ieee_fullname}
\bibliography{refs}
}


\clearpage

\onecolumn
\begin{center}
{\Large\bf{Boundless: Generative Adversarial Networks for Image Extension - Supplementary Material}\\
}
\end{center}

\maketitle

\begin{multicols}{2}



\section{Network Training and Architecture Details}

\subsection{Generator Network $\bm{G}$}

\begin{center}
\small
\setlength{\tabcolsep}{4pt}
\centering
\begin{tabular}{|c|c|c|c|c|c|c|c|}
\hline
Layer ID&Type&Act.&K&S&D&Out&Skip \\
\hline \hline
1 & Gated Conv  & ELU\cite{clevert2015fast} & 5 & 1 & 1 & 32 & None\\
\hline
2 & Gated Conv  & ELU & 3 & 2 & 1 & 64 & None\\
\hline
3 & Gated Conv  & ELU & 3 & 1 & 1 & 64 & None\\
\hline
4 & Gated Conv  & ELU & 3 & 2 & 1 & 128 & None\\
\hline
5 & Gated Conv  & ELU & 3 & 1 & 1 & 128 & None\\
\hline
6 & Gated Conv  & ELU & 3 & 1 & 1 & 128 & None\\
\hline
7 & Gated Conv  & ELU & 3 & 1 & 2 & 128 & None\\
\hline
8 & Gated Conv  & ELU & 3 & 1 & 4 & 128 & None\\
\hline
9 & Gated Conv  & ELU & 3 & 1 & 8 & 128 & None\\
\hline
10 & Gated Conv  & ELU & 3 & 1 & 16 & 128 & None\\
\hline
11 & Gated Conv  & ELU & 3 & 1 & 1 & 128 & 5\\
\hline
12 & Gated Conv  & ELU & 3 & 1 & 1 & 128 & 4\\
\hline
13 & Resize (2x)  & n/a & n/a & n/a & n/a & n/a & n/a\\
\hline
14 & Gated Conv  & ELU & 3 & 1 & 1 & 64 & 3\\
\hline
15 & Gated Conv  & ELU & 3 & 1 & 1 & 64 & 2\\
\hline
16 & Resize (2x)  & n/a & n/a & 1 & n/a & n/a & n/a\\
\hline
17 & Gated Conv  & ELU & 3 & 1 & 1 & 32 & 1\\
\hline
18 & Gated Conv  & ELU & 3 & 1 & 1 & 16 & None\\
\hline
19 & Conv  & None & 3 & 1 & 1 & 3 & None\\
\hline
20 & Clip  & n/a & n/a & n/a & n/a & n/a & n/a\\
\hline
\end{tabular}
\newline
\captionof{table}{The generator architecture. Act. stands for activation type, K stands for kernel size, S for stride, D for dilation, Out for number of channels in convolutional layers and number of units in fully connected units, and Skip represents the layer-id which is concatenated into the output of the given layer. All resize operations use  bilinear interpolation. In the Generator, all convolutional layers use 'Same' padding. }
\end{center}
\subsection{Discriminator Network $\bm{D}$}

The discriminator applies spectral normalization \cite{miyato2018spectral} at all layers, and consists of the the common tower ($\bm{D_N}$, Table \ref{arch:common_tower}), which feeds into  the non-conditional branch ($\bm{f_N}$, Table \ref{arch:main_branch}) and projection discriminator branch ($\bm{f_C}$, Table \ref{arch:cond_branch}). These two branches produce scalars, which are then summed to produce a single network output. We invite the reader to see Section 3 of the main paper for more in depth discussion of the model. 

The scalar outputs of the main and projection discriminator are summed and passed to the adversarial loss.

\begin{center}
\centerline{\textbf{Common Tower} $\bm{D_N}$}
\vspace{2mm}
\small
\setlength{\tabcolsep}{4pt}
\centering
\begin{tabular}{|c|c|c|c|c|c|c|}
\hline 
Layer ID&Type&Act.&K&S&Padding&Out Size \\
\hline \hline
1 & Conv  & LeakyReLU\cite{maas2013rectifier} & 5 & 2  & Same & 64\\
\hline
2 &  Conv  & LeakyReLU & 5 & 2  & Same & 128\\
\hline
3 &  Conv  & LeakyReLU& 5 & 2 &  Same & 256\\
\hline
4 &  Conv  & LeakyReLU & 5 & 2 & Same & 256\\
\hline
5 &  Conv  & LeakyReLU & 5 & 2 & Same & 256\\
\hline
6 &  Conv  & LeakyReLU & 5 & 2 & Same & 256\\
\hline
7 &  Conv  & LeakyReLU & 5 & 1 & Valid & 256\\
\hline
8 &  Flatten  & n/a & n/a & n/a & n/a & n/a\\
\hline
\end{tabular}
\newline
\captionof{table}{The base of the discriminator. It takes  generated and ground truth images as input. Act. stands for activation type, K stands for kernel size, S for stride, Out for number of channels in convolutional layers and number of units in fully connected units. }
\label{arch:common_tower}
\end{center}

\begin{center}
\centerline{\textbf{Non-Conditional Branch} $\bm{f_N}$}
\vspace{2mm}
\small
\setlength{\tabcolsep}{4pt}
\centering
\begin{tabular}{|c|c|c|c|c|c|c|}
\hline 
Layer ID&Type&Act.&Out Size \\
\hline \hline
1 & Fully Connected No Bias  & None & 1\\
\hline
\end{tabular}
\newline
\captionof{table}{The non-conditional branch of the discriminator, taking the common tower from Table 2 as input and outputting a single scalar value. Act. stands for activation type.  }
\label{arch:main_branch}
\end{center}

\begin{center}
\centerline{\textbf{Projection Discriminator Branch} $\bm{f_C}$ }
\vspace{2mm}
\small
\setlength{\tabcolsep}{4pt}
\centering
\begin{tabular}{|c|c|c|c|c|c|c|}
\hline 
Layer ID&Type&Act.&Out Size \\
\hline \hline
1 & Normalize  & None  & 1000\\
\hline
2 & Fully Connected No Bias  & None  & 256\\
\hline
3 & Inner Product w/Common Tower  & None  & 1\\
\hline
\end{tabular}
\newline
\captionof{table}{The projection discriminator \cite{miyato2018cgans} branch of the network. The input is logits of a pretrained classification network, for which we used an InceptionV3 \cite{szegedy2016rethinking} network trained on ImageNet \cite{deng2009imagenet}. The output is a single scalar, which is summed with the output of the non-conditional branch and passed to the hinge loss.}
\label{arch:cond_branch}
\end{center}

\subsection{Training details:}

We take the training set of Places365-Challenge dataset \cite{zhou2017places}, select the top 50 classes by number of samples, and create a holdout validation set from this. This creates about 39,000 training and 930 test samples per class, for a total training set size of 1,953,624 and testing set size of 46376. The classes selected are:

\begin{itemize}
\setlength\itemsep{-0.5em}
\item amusement park
\item aquarium
\item athletic field
\item baseball field
\item bathroom
\item beach
\item bridge
\item building facade
\item car interior
\item church - indoor
\item church - outdoor
\item cliff
\item coast
\item corridor
\item dining room
\item embassy
\item forest
\item forest path
\item golf course
\item harbor
\item highway
\item industrial area
\item lagoon
\item lake
\item lighthouse
\item living room
\item lobby
\item mansion
\item mountain
\item ocean
\item office building
\item palace
\item parking lot
\item pier
\item pond
\item porch
\item railroad track
\item rainforest
\item river
\item skyscraper
\item stadium
\item staircase
\item swamp
\item swimming hole
\item swimming pool
\item train station
\item underwater
\item valley
\item vegetable garden
\item water park
\end{itemize}

Before passing the training image into the generator we resize the image to 257 x 257, and also concatenate the mask channel. The mask size is randomly sampled from a uniform distribution, which is the target size plus/minus 4 pixels, so the model doesn't overfit to a specific mask size. 

Following the code of DeepFill \cite{yu2018generative}, we concatenate a channel of 1's to the input of the generator. This enables the generator to see the edge of the image after 0 padding the inputs, although we do not verify this in this work. 

We take generator and discriminator steps in a 1:1 ratio, with the steps executed jointly. 

Please see Section 3 of the main paper for more discussion of the loss and optimizer.

\section{Qualitative Results}

We show additional samples from on the 25\%, 50\%, and 75\%  mask image extension experiments, and refer the reader to Figures \ref{fig:quarter}, \ref{fig:half}, and \ref{fig:three_quarter}. We also show additional results from in-painting experiment in Figure \ref{fig:inpainting} and more panorama results in Figure \ref{fig:pano}.We also demonstrate the suitability of our method on freeform masks in Figure \ref{fig:freeform}.

\begin{center}
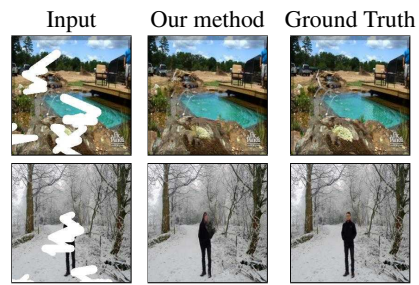

\newcommand{\figwidth}{.09\textwidth}
\newcommand{\shiftleft}{\hspace{-20pt}}
\newcommand{\figurefile}[2]{\frame{\includegraphics[width=\figwidth]{figs/freeform_inpainting/#1/#2}}}
\newcommand{\onerow}[1]{
\figurefile{#1}{input.jpg} & 
\figurefile{#1}{ffm.jpg} & 
\figurefile{#1}{gt.jpg} 
}
\setlength{\tabcolsep}{1.2mm}
\begin{tabular}{@{\hskip 15mm}c c c}
 {\small Input}  & {\small Our method}  &  {\small Ground Truth}\\
\onerow{example_1}\\
\onerow{example_2}\\
\end{tabular}
\captionof{figure}{\small Results on freeform masks. }
\label{fig:freeform}
\end{center}

\begin{figure*}
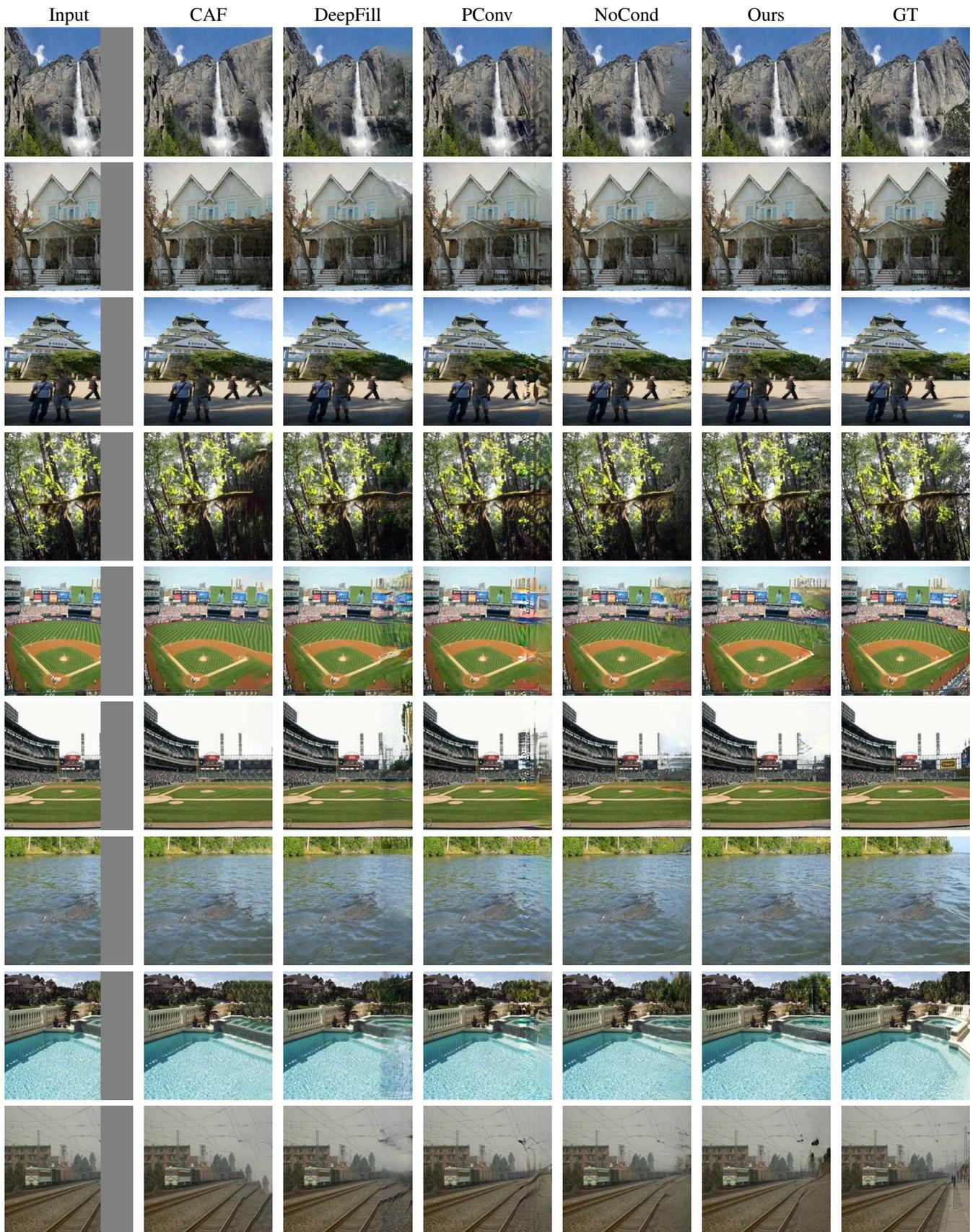

\newcommand{\figwidth}{.135\textwidth}
\newcommand{\shiftleft}{\hspace{-6pt}}
\newcommand{\figurefile}[2]{\shiftleft\includegraphics[width=\figwidth]{figs/quarter_uncrop/#1/#2}}
\newcommand{\onerow}[1]{\figurefile{#1}{input.jpg} & \figurefile{#1}{caf.jpg} & \figurefile{#1}{df.jpg} & \figurefile{#1}{pcnv.jpg} & \figurefile{#1}{uncond.jpg} & \figurefile{#1}{cond.jpg} & \figurefile{#1}{gt.jpg}}
\begin{tabular}{ccccccc}
\shiftleft Input & \shiftleft CAF & \shiftleft DeepFill & \shiftleft PConv & \shiftleft NoCond & \shiftleft Ours & \shiftleft GT\\
\onerow{example_1}\\
\onerow{example_2}\\
\onerow{example_3}\\
\onerow{example_5}\\
\onerow{example_6}\\
\onerow{example_7}\\
\onerow{example_8}\\
\onerow{example_9}\\
\onerow{example_10}
\end{tabular}
\caption{Extending images from masks which are 25\% of the image width. We note that edges and structure are better defined in our method. For instance, edge of the roof in the second row.}
\label{fig:quarter}
\end{figure*}

\begin{figure*}
\newcommand{\figwidth}{.135\textwidth}
\newcommand{\shiftleft}{\hspace{-6pt}}
\newcommand{\figurefile}[2]{\shiftleft\includegraphics[width=\figwidth]{figs/half_uncrop/#1/#2}}
\newcommand{\onerow}[1]{\figurefile{#1}{input.jpg} & \figurefile{#1}{caf.jpg} & \figurefile{#1}{df.jpg} & \figurefile{#1}{pcnv.jpg} & \figurefile{#1}{uncond.jpg} & \figurefile{#1}{cond.jpg} & \figurefile{#1}{gt.jpg}}
\begin{tabular}{ccccccc}
\shiftleft Input & \shiftleft CAF & \shiftleft DeepFill & \shiftleft PConv & \shiftleft NoCond & \shiftleft Ours & \shiftleft GT\\
\onerow{example_1}\\
\onerow{example_2}\\
\onerow{example_3}\\
\onerow{example_4}\\
\onerow{example_5}\\
\onerow{example_7}\\
\onerow{example_8}\\
\onerow{example_9}\\
\onerow{example_10}
\end{tabular}
\caption{Extending images from masks which are 50\% of the image width.}
\label{fig:half}
\end{figure*}

\begin{figure*}
\newcommand{\figwidth}{.135\textwidth}
\newcommand{\shiftleft}{\hspace{-6pt}}
\newcommand{\figurefile}[2]{\shiftleft\includegraphics[width=\figwidth]{figs/three_quarter_uncrop/#1/#2}}
\newcommand{\onerow}[1]{\figurefile{#1}{input.jpg} & \figurefile{#1}{caf.jpg} & \figurefile{#1}{df.jpg} & \figurefile{#1}{pcnv.jpg} & \figurefile{#1}{uncond.jpg} & \figurefile{#1}{cond.jpg} & \figurefile{#1}{gt.jpg}}
\begin{tabular}{ccccccc}
\shiftleft Input & \shiftleft CAF & \shiftleft DeepFill & \shiftleft PConv & \shiftleft NoCond & \shiftleft Ours & \shiftleft GT\\
\onerow{example_1}\\
\onerow{example_2}\\
\onerow{example_3}\\
\onerow{example_4}\\
\onerow{example_5}\\
\onerow{example_7}\\
\onerow{example_8}\\
\onerow{example_9}\\
\onerow{example_10}
\end{tabular}
\caption{Extending images from masks which are 75\% of the image width.}
\label{fig:three_quarter}
\end{figure*}

\begin{figure*}
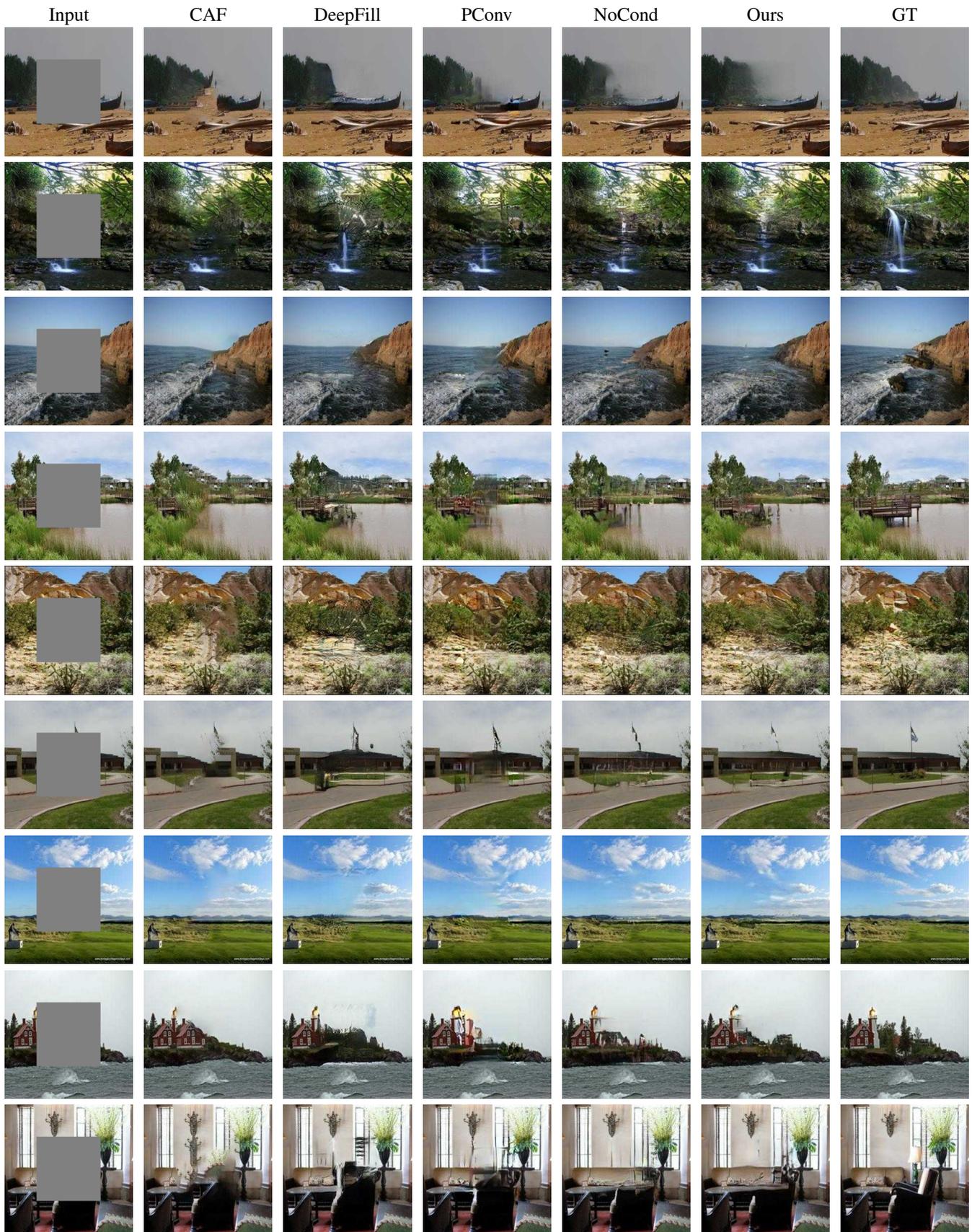

\newcommand{\figwidth}{.135\textwidth}
\newcommand{\shiftleft}{\hspace{-6pt}}
\newcommand{\figurefile}[2]{\shiftleft\includegraphics[width=\figwidth]{figs/inpainting/#1/#2}}
\newcommand{\onerow}[1]{\figurefile{#1}{input.jpg} & \figurefile{#1}{caf.jpg} & \figurefile{#1}{df.jpg} & \figurefile{#1}{pcnv.jpg} & \figurefile{#1}{uncond.jpg} & \figurefile{#1}{cond.jpg} & \figurefile{#1}{gt.jpg}}
\begin{tabular}{ccccccc}
\shiftleft Input & \shiftleft CAF & \shiftleft DeepFill & \shiftleft PConv & \shiftleft NoCond & \shiftleft Ours & \shiftleft GT\\
\onerow{example_1}\\
\onerow{example_2}\\
\onerow{example_3}\\
\onerow{example_4}\\
\onerow{example_6}\\
\onerow{example_7}\\
\onerow{example_8}\\
\onerow{example_9}\\
\onerow{example_10}
\end{tabular}
\caption{Center Inpainting.}
\label{fig:inpainting}
\end{figure*}

\begin{figure*}
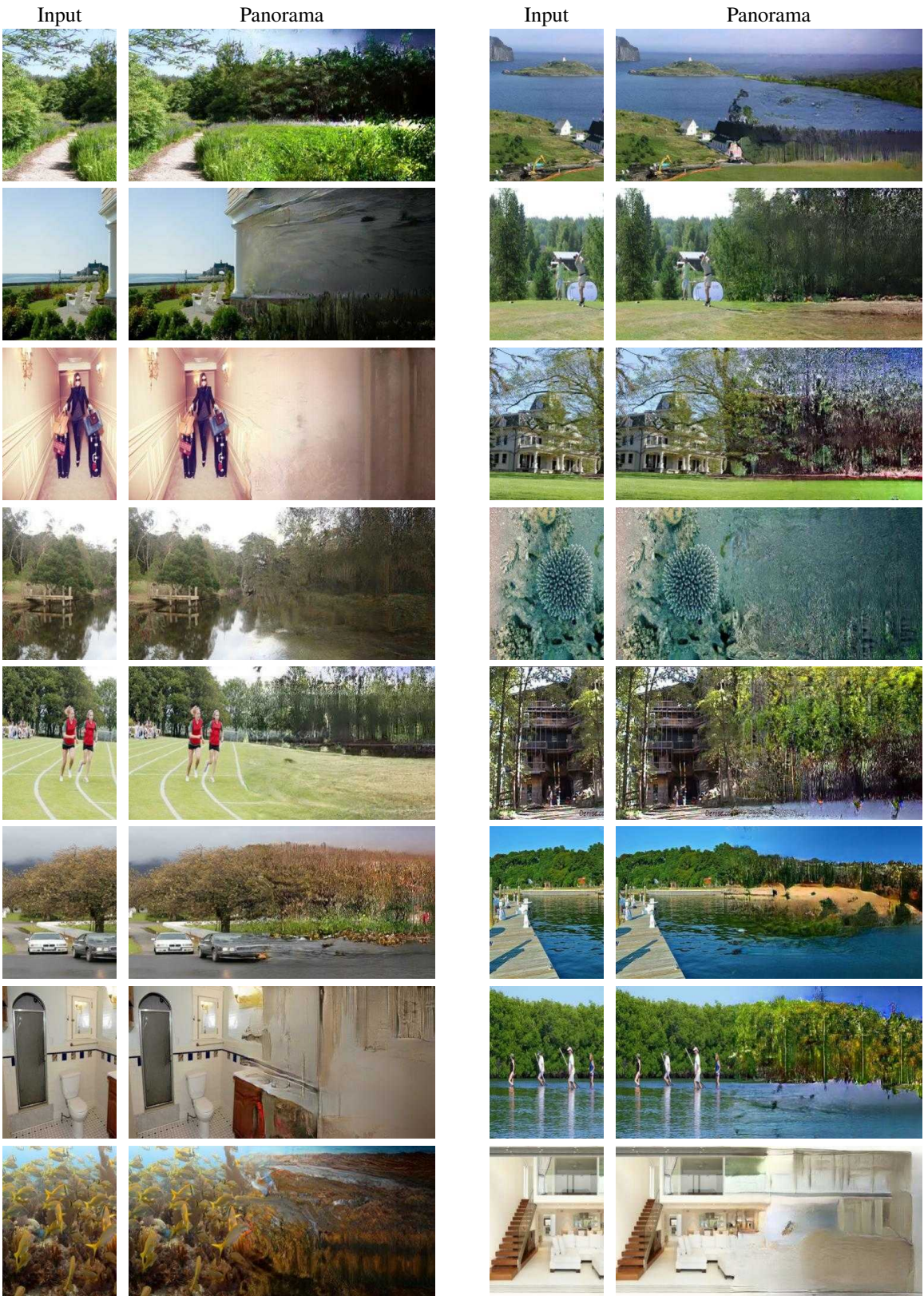

\newcommand{\figwidth}{.135\textwidth}
\newcommand{\figheight}{72pt}
\newcommand{\panowidth}{0.675\textwidth}
\newcommand{\shiftleft}{\hspace{-6pt}}
\newcommand{\figurefile}[2]{\shiftleft\includegraphics[height=\figheight]{figs/pano/#1/#2}}
\newcommand{\onepano}[1]{\figurefile{#1}{input.jpg} & \figurefile{#1}{cond.jpg}}
\newcommand{\onerow}[2]{\onepano{#1} & \onepano{#2}}
\centering
\begin{tabular}{cc@{\hspace{32pt}}cc}
\shiftleft Input & \shiftleft Panorama & \shiftleft Input & \shiftleft Panorama \\
\onerow{example_1}{example_2}\\
\onerow{example_3}{example_4}\\
\onerow{example_5}{example_6}\\
\onerow{example_7}{example_8}\\
\onerow{example_9}{example_10}\\
\onerow{example_11}{example_12}\\
\onerow{example_13}{example_14}\\
\onerow{example_15}{example_16}\\
\end{tabular}
\caption{Additional panorama results}
\label{fig:pano}
\end{figure*}

\section{Exploring the Space of Plausible Extensions} 

We invite the reader to view the accompanying video derived from a sample from the YouTube8m dataset \cite{abu2016youtube} at \url{https://drive.google.com/file/d/1x6FCYPmoqSuCdeLJTD0UpQ_MQhBPv7_e/view?usp=sharing}.  Please refer to the main paper for details on how it was created. We encourage the reader to pause the video at arbitrary frames to see how the model produces different plausible completions as the result of tiny perturbations of the original frame.

\section{Failure Cases}

\begin{figure*}
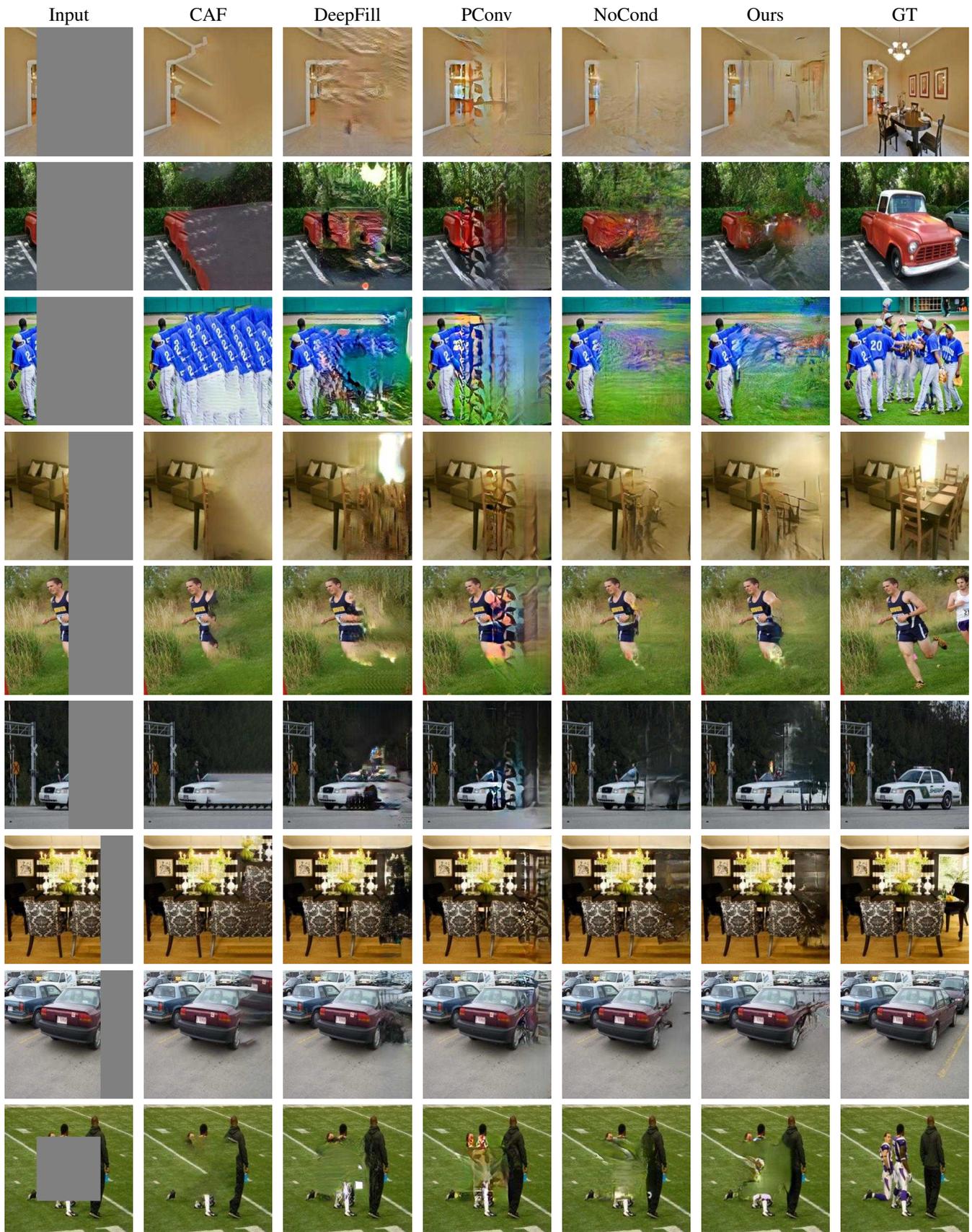

\newcommand{\figwidth}{.135\textwidth}
\newcommand{\shiftleft}{\hspace{-6pt}}
\newcommand{\figurefile}[2]{\shiftleft\includegraphics[width=\figwidth]{figs/failure/#1/#2}}
\newcommand{\onerow}[1]{\figurefile{#1}{input.jpg} & \figurefile{#1}{caf.jpg} & \figurefile{#1}{df.jpg} & \figurefile{#1}{pcnv.jpg} & \figurefile{#1}{uncond.jpg} & \figurefile{#1}{cond.jpg} & \figurefile{#1}{gt.jpg}}
\begin{tabular}{ccccccc}
\shiftleft Input & \shiftleft CAF & \shiftleft DeepFill & \shiftleft PConv & \shiftleft NoCond & \shiftleft Ours & \shiftleft GT\\
\onerow{example_1}\\
\onerow{example_2}\\
\onerow{example_3}\\
\onerow{example_4}\\
\onerow{example_5}\\
\onerow{example_6}\\
\onerow{example_7}\\
\onerow{example_8}\\
\onerow{example_9}\\
\end{tabular}
\caption{Failure cases. The network struggles with objects; especially cars, humans, and furniture.}
\label{fig:epic_gan_fail}
\end{figure*}

In Figure \ref{fig:epic_gan_fail} we examine some of the failure modes of our image extension model. We note that our model is much better at textures than objects; for example vehicles, people, and furniture are challenging for the model. Addressing this is left to future work. 

\end{multicols}
\end{document}


\title{Conditional Generative Adversarial Networks for Image Extension - Supplementary Material}

\author{First Author\\
Institution1\\
Institution1 address\\
{\tt\small firstauthor@i1.org}
\and
Second Author\\
Institution2\\
First line of institution2 address\\
{\tt\small secondauthor@i2.org}
}

\maketitle

blah bloo

{\small
\bibliographystyle{ieee}
\bibliography{refs}
}